\documentclass[10pt,twocolumn,letterpaper]{article}

\usepackage[pagenumbers]{cvpr} 

\definecolor{cvprblue}{rgb}{0.21,0.49,0.74}
\usepackage[pagebackref,breaklinks,colorlinks,allcolors=cvprblue]{hyperref}

\def\paperID{} 
\def\confName{CVPR}
\def\confYear{2026}

\title{A Trajectory-Driven Spatio-Temporal Refinement Solution for \\ CVPR 2026 8th UG2+ Challenge Track 3: DOST}

\author{Hongzhen Li \quad Miao Yu \quad Leilei Cao\thanks{Corresponding author}  \quad Youwei Pan \quad Yingfang Zhu \quad Fengjie Zhu\\
TEX AI, Transsion Holdings\\
{\tt\small lihongzhen0724@gmail.com, leilei.cao@transsion.com}
}

\begin{document}
\maketitle
\begin{abstract}
In this work, we present our solution for the 8th UG2+ Challenge (CVPR 2026) Track 3: Dynamic Object Segmentation in Turbulence (DOST). Our method is built upon the strong baseline framework Segment Any Motion (SegAnyMo), which provides powerful mask generation and motion tracking capabilities. To further boost the segmentation performance under severe atmospheric distortions, we propose two key improvements. First, we employ a data-centric domain adaptation strategy. We significantly expand our training data by incorporating selected sequences from the DAVIS dataset alongside a subset of the DOST dataset, and apply simulated atmospheric fluctuation degradations to enhance the model's robustness against complex geometric distortions. Second, we introduce a spatio-temporal post-processing module. This refinement step effectively removes persistent boundary-connected false foregrounds and short-lived fragmented noise, while strictly preserving genuine small targets and maintaining original individual labels across frames. With these combined strategies, our proposed method ranks the 2st place in the challenge.
\end{abstract}    
\section{Introduction}
\label{sec:intro}

Moving Object Segmentation (MOS) is a fundamental task in computer vision, serving as a cornerstone for numerous downstream applications such as autonomous driving, video surveillance, and robotic navigation. While recent advancements in Video Object Segmentation (VOS) have achieved remarkable performance on standard benchmarks, these models often assume ideal imaging conditions. The 8th UG2+ Challenge (CVPR 2026) Track 3: Dynamic Object Segmentation in Turbulence (DOST) introduces an extremely challenging, real-world scenario where videos are severely downgraded by atmospheric turbulence.

The inherent characteristics of the DOST \cite{Pont-Tuset_arXiv_2017} dataset present unprecedented obstacles. Atmospheric turbulence introduces spatially and time-varying distortions, severe non-rigid geometric deformations, and high-frequency temporal jitter. These complex degradations fundamentally violate the "brightness constancy" assumption that traditional optical flow-based methods strictly rely upon. Consequently, conventional short-term motion estimation and dense pixel-matching algorithms often collapse in turbulent environments, leading to fragmented predictions, severe boundary drifting, and temporally inconsistent masks.

To effectively tackle these issues, we abandon complex end-to-end network redesigns and instead propose a highly robust pipeline built upon the foundation model Segment Any Motion (SegAnyMo) \cite{huang2025segment}. Our motivation is twofold. First, instead of relying on adjacent-frame dense flow, SegAnyMo extracts long-range point trajectories across the entire video sequence. This long-term temporal perspective acts as a natural low-pass filter, effectively mitigating the localized structural collapse and high-frequency random jitter caused by heat waves. Second, by employing a motion-semantic decoupled design, it utilizes the filtered dynamic trajectories to prompt the frozen SAM 2 \cite{ravi2025sam} model, enabling powerful zero-shot mask generation even under severe geometric distortions.

Despite the robustness of the baseline, a significant domain gap remains between clean tracking datasets and the turbulent DOST scenarios. To bridge this gap, we adopt a data-centric domain adaptation strategy. We empirically found that applying our physics-inspired turbulence simulation to high-fidelity datasets like DAVIS \cite{qin2024unsupervised} provides the most stable motion priors, allowing the network to isolate real object movements from atmospheric jitter without catastrophic forgetting. Furthermore, recognizing the persistent boundary artifacts and short-lived noise specific to turbulent videos, we design a spatio-temporal post-processing module to refine the final masks.

\section{Proposed Method}
\label{sec:method}

To address the severe geometric distortions and temporal jitter caused by atmospheric turbulence, we propose a robust and clean pipeline. Instead of relying on vulnerable optical flow methods, our framework leverages long-range point trajectories combined with physics-inspired domain adaptation and targeted spatio-temporal refinement.

\subsection{Baseline: Segment Any Motion}
\label{subsec:baseline}

Conventional moving object segmentation heavily relies on optical flow, which frequently collapses under turbulence due to the violation of the brightness constancy assumption. To overcome this limitation, we build our pipeline on the \textbf{Segment Any Motion (SegAnyMo)} framework. 

The core process of our baseline consists of two main stages. First, rather than computing dense pixel matching between adjacent frames, the model extracts long-range point trajectories across the entire video sequence. This long-term temporal view acts as a natural low-pass filter, which effectively smooths out the high-frequency random vibrations caused by heat waves. Second, a Spatio-Temporal Trajectory Attention module isolates genuine moving points from background movements. These robust trajectories are then used as sparse prompts to guide the frozen SAM 2 model for pixel-level mask generation. This motion-semantic decoupled design ensures high-quality zero-shot segmentation without requiring dense turbulent annotations during pre-training.

\subsection{Data-Centric Domain Adaptation}
\label{subsec:data}

A noticeable domain gap exists between standard clean video datasets and turbulent scenarios. To bridge this gap, we implement a data-centric training strategy featuring tailored data filtering and a physics-inspired turbulence simulator.

\vspace{1mm}
\noindent \textbf{Data Selection Criteria.} While large-scale video object segmentation datasets like YouTube-VOS, SegTrackv2, and FBMS59 are widely used, we empirically find them sub-optimal for turbulence adaptation. Older datasets like SegTrackv2 suffer from low-resolution and compression artifacts, while FBMS59 provides only temporally sparse annotations. Applying synthetic turbulence to these low-fidelity videos catastrophically disrupts the trajectory tracking module. Therefore, we strictly restrict our training data to the high-fidelity \textbf{DAVIS} dataset, which provides dense temporal annotations and continuous motion patterns. This serves as a clean canvas for the tracking baseline to learn stable motion priors.

\vspace{1mm}
\noindent \textbf{Turbulence Simulation.} To embed turbulence-resistant priors, we design a simulator to mimic atmospheric heat waves. We first generate random spatial noise matrices $\mathcal{N}_x, \mathcal{N}_y \sim \mathcal{N}(0, 1)$ and apply a spatial low-pass filter using a Gaussian kernel with a standard deviation $\sigma_{wave}$ to model the scale of turbulence wave cells:
\begin{equation}
    S_x = G_{\sigma_{wave}} * \mathcal{N}_x, \quad S_y = G_{\sigma_{wave}} * \mathcal{N}_y
\end{equation}
To enforce temporal continuity, the offset fields are smoothed across adjacent frames using a first-order Exponential Moving Average (EMA) with a factor $\alpha$:
\begin{equation}
    T_t = \alpha S_t + (1 - \alpha) T_{t-1}
\end{equation}
The smoothed field is scaled by an intensity factor $\lambda$ to geometrically warp the clean frame $I_t$ via bilinear interpolation:
\begin{equation}
    \tilde{I}_t(x, y) = I_t(x + \lambda T_{t,x}, y + \lambda T_{t,y})
\end{equation}
Finally, a mild Gaussian blur is applied to simulate optical scattering. Empirically, extreme degradations impede network convergence. Thus, we adopt a mild, stochastic training strategy to preserve tracking stability while gaining turbulence robustness.

\subsection{Spatio-Temporal Post-Processing}
\label{subsec:post}

The masks produced by the baseline often contain false positives near the borders and isolated fragmentation caused by severe turbulence. We introduce a spatio-temporal post-processing module to filter these artifacts while preserving multi-object individual labels.

\vspace{1mm}
\noindent \textbf{Persistent Boundary Artifact Removal.} Atmospheric distortions frequently cause background pixels near the video borders to be misclassified as foreground. We define a boundary zone within a specific pixel width $W_b$. If a predicted foreground component continuously connects to the frame boundaries and its ratio of boundary-touching pixels exceeds a threshold $\gamma$, it is classified as a persistent false foreground and eliminated. 

\vspace{1mm}
\noindent \textbf{Temporal Consistency Filtering.} Turbulence also introduces short-lived, small-area mask fragments. For any foreground component with a spatial area smaller than $A_{min}$, we evaluate its temporal persistence within a sliding window $V_t$. If the component fails to maintain a stable spatio-temporal intersection-over-union (IoU) across consecutive frames, it is discarded as noise. Conversely, genuine small targets that consistently appear across frames are strictly preserved. Crucially, this refinement operates independently on each individual object label, avoiding the unintended merging of distinct target instances.
\begin{figure*}[t] 
  \centering
  \includegraphics[width=0.95\linewidth]{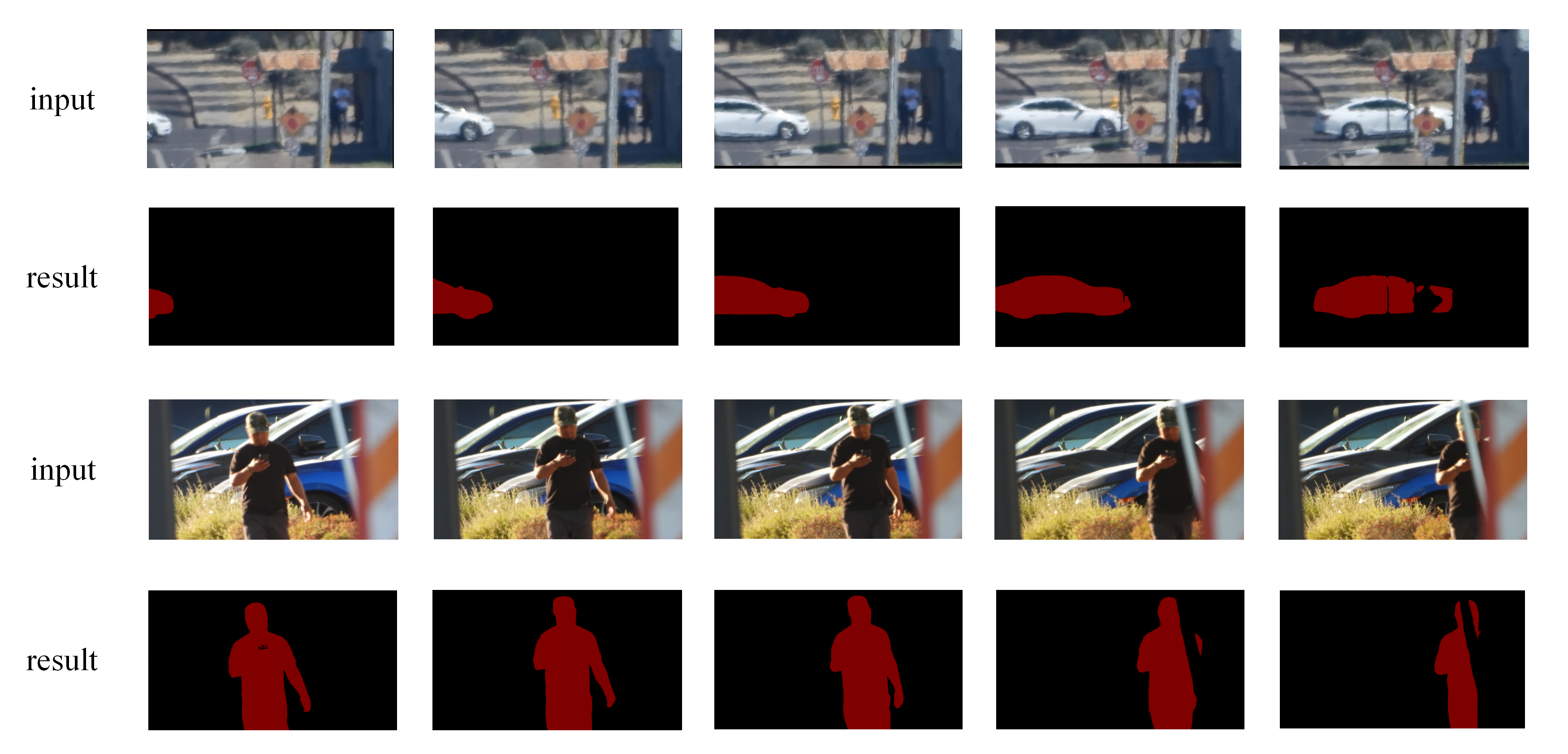}
  \caption{Qualitative results on the DOST testing set}
  \label{fig:results}
\end{figure*}

\section{Experiments}
\label{sec:experiments}

In this section, we present the implementation details, empirical observations during training, and the final quantitative results on the DOST dataset.

\subsection{Implementation Details}
\label{subsec:exp_details}

Our proposed pipeline is implemented in PyTorch and all experiments are conducted on a single NVIDIA H100 (80GB) GPU. 

\vspace{1mm}
\noindent \textbf{Network Configuration.} We adopt the \textit{traj\_oa\_depth} architecture from the SegAnyMo framework. To provide robust inputs for the motion-semantic decoder, we utilize \textit{BootSTAP} \cite{doersch2024bootstap} for long-range trajectory extraction and \textit{Depth-Anything-V2} \cite{yang2024depth} for monocular depth estimation. The DINOv2 \cite{oquab2023dinov2} module is also enabled to extract high-level semantic features for spatial-temporal attention coupling.

\vspace{1mm}
\noindent \textbf{Training Hyperparameters.} The model is initialized with the official pre-trained weights and fine-tuned for 40 epochs. The batch size is set to 1. We optimize the network using a learning rate of $1 \times 10^{-4}$ and a weight decay of $1 \times 10^{-4}$. During inference, the iterative prompting module of SAM 2 is utilized to generate the final dense masks.

\vspace{1mm}
\noindent \textbf{Post-Processing Settings.} According to our refinement strategy, we set the boundary width $W_b = 10$ pixels to eliminate border artifacts. To filter temporal noise, the minimum spatial area is set to 10 pixels, and temporal persistence is evaluated across a sliding window of 3 frames.


\subsection{Training Strategy and Dataset Observations}
\label{subsec:dataset_obs}

During our preliminary experiments, we attempted to fine-tune the pre-trained model using the complete DOST dataset combined with large-scale datasets such as YouTube-VOS (YTVOS18m) \cite{xu2018youtube}. However, we observed a rapid and severe performance degradation (catastrophic forgetting). 

We hypothesize that this degradation mainly stems from a severe domain gap. The extreme geometric distortions and random temporal jitters in the full DOST dataset likely disrupt the pre-trained feature manifold. Furthermore, incorporating vast sequences from YTVOS18m might introduce conflicting low-resolution artifacts and complex camera motions, which could potentially confuse the long-range trajectory tracker.

Consequently, to maintain the stability of the foundation model, we strictly curbed the training data. We constructed a specialized subset by combining clean sequences from the high-fidelity DAVIS dataset with carefully selected, moderately degraded sequences from the DOST dataset. This strategy preserved the pre-trained tracking capabilities.


\subsection{Evaluation Metrics}
\label{subsec:metrics}

Following the standard evaluation protocol of the 8th UG2+ Challenge, we report the mean Intersection over Union (mIoU) for region similarity and the mean Dice coefficient (mDice) for contour accuracy.

\begin{table}[t]
  \caption{Quantitative results on the DOST testing set.}
  \label{tab:results}
  \centering
  \begin{tabular}{lcc}
    \toprule
    Method & mIoU & mDice \\
    \midrule
    SegAnyMo (Baseline) & 0.6392 & 0.6953 \\
    \textbf{Ours} & \textbf{0.7249} & \textbf{0.7783} \\
    \bottomrule
  \end{tabular}
\end{table}

\subsection{Quantitative and Qualitative Results}
\label{subsec:results}

Our final submission integrates the data-centric domain adaptation and the spatio-temporal post-processing module. As shown in \cref{tab:results} and \cref{fig:results}, our proposed pipeline significantly improves the segmentation robustness under severe atmospheric turbulence compared to the baseline. The results demonstrate the effectiveness of our data selection criteria and post-processing refinement in handling highly distorted non-rigid targets.
\section{Conclusion}
\label{sec:conclusion}

In this report, we presented our solution for the 8th UG2+ Challenge Track 3: Dynamic Object Segmentation in Turbulence. To overcome the extreme geometric distortions and temporal jitter inherent in turbulent environments, we proposed a robust method built upon the SegAnyMo framework. By leveraging long-range point trajectories rather than adjacent-frame dense flow, the baseline gains inherent resistance to high-frequency turbulence. Furthermore, we implemented a data-centric domain adaptation strategy that incorporates a  turbulence simulator on high-fidelity data. Finally, we designed a spatio-temporal post-processing module to eliminate persistent boundary artifacts and temporal noise. Experimental results demonstrate that our method improves segmentation accuracy under severe atmospheric degradation.
{
    \small
    \bibliographystyle{ieeenat_fullname}
    \bibliography{main}
}


\end{document}